\documentclass{article}
\usepackage{ijcai19}

\usepackage{times}
\usepackage{soul}
\usepackage{url}
\usepackage[hidelinks]{hyperref}
\usepackage[utf8]{inputenc}
\usepackage[small]{caption}
\usepackage{graphicx}
\usepackage{amsmath}
\usepackage{booktabs}
\usepackage{algorithm}
\usepackage{algorithmic}
\usepackage[list=false]{subcaption}
\usepackage{amssymb}
\usepackage{tablefootnote}
\usepackage{siunitx} 
\usepackage{multicol}				
\usepackage{multirow}
\usepackage{hhline}

\usepackage{mathtools}  
    \usepackage{xfrac}   

\usepackage{arydshln}

\graphicspath{{./images/}}

\newcommand{\Expec}{\displaystyle \mathbb{E}}
\newcommand{\Real}{\mathbb{R}}

\def\PDQN*{P\nobreakdash-DQN}
\def\SPDQN*{SP\nobreakdash-DQN}
\def\MPDQN*{MP\nobreakdash-DQN}
\def\QPAMDP*{Q\nobreakdash-PAMDP}
\def\PADDPG*{PA\nobreakdash-DDPG}

\newcommand{\citet}[1]
{\citeauthor{#1}~\shortcite{#1}}
\newcommand{\citep}{\cite}

\makeatletter
\def\adl@drawiv#1#2#3{%
	\hskip.5\tabcolsep
	\xleaders#3{#2.5\@tempdimb #1{1}#2.5\@tempdimb}%
	#2\z@ plus1fil minus1fil\relax
	\hskip.5\tabcolsep}
\newcommand{\cdashlinelr}[1]{%
	\noalign{\vskip\aboverulesep
		\global\let\@dashdrawstore\adl@draw
		\global\let\adl@draw\adl@drawiv}
	\cdashline{#1}
	\noalign{\global\let\adl@draw\@dashdrawstore
		\vskip\belowrulesep}}
\makeatother

\newrobustcmd{\B}{\fontseries{b}\selectfont}
\renewrobustcmd{\boldmath}{}

\title{Multi-Pass Q-Networks for Deep Reinforcement Learning with~\\Parameterised Action Spaces}

\author{
Craig J. Bester$^1$\and
Steven D. James$^1$\and
George D. Konidaris$^2$\\
\affiliations
$^1$University of the Witwatersrand, Johannesburg\\
$^2$Brown University, Providence RI\\
}

\begin{document}

\maketitle

\begin{abstract}
Parameterised actions in reinforcement learning are composed of discrete actions with continuous action-parameters. This provides a framework for solving complex domains that require combining high-level actions with flexible control. The recent \PDQN* algorithm extends deep Q-networks to learn over such action spaces. However, it treats all action-parameters as a single joint input to the Q-network, invalidating its theoretical foundations. We analyse the issues with this approach and propose a novel method---multi-pass deep Q-networks, or \MPDQN*---to address them. We empirically demonstrate that \MPDQN* significantly outperforms \PDQN* and other previous algorithms in terms of data efficiency and converged policy performance on the Platform, Robot Soccer Goal, and Half Field Offense domains.
\end{abstract}

\section{Introduction}

Reinforcement learning (RL) and deep RL in particular have demonstrated remarkable success in solving tasks that require either discrete actions, such as Atari \citep{mnih2015}, or continuous actions, such as robot control \citep{schulman2015,lillicrap2015}. Reinforcement learning with parameterised actions \citep{masson2016} that combine discrete actions with continuous action-parameters has recently 
emerged as an additional setting of interest,  
allowing agents to learn flexible behavior in tasks such as 2D robot soccer
\citep{hausknecht2016,hussein2018}, simulated human-robot interaction \citep{khamassi2017}, and terrain-adaptive bipedal and quadrupedal locomotion \citep{peng2016}.


There are two main approaches to learning with parameterised actions: alternate between optimising the discrete actions and continuous action-parameters separately \citep{masson2016,khamassi2017}, or collapse the parameterised action space into a continuous one \citep{hausknecht2016}. Both of these approaches fail to fully exploit the structure present in parameterised action problems. The former does not share information between the action and action-parameter policies, while the latter does not take into account which action-parameter is associated with which action, or even which discrete action is executed by the agent. More recently, \citet{xiong2018} introduced \PDQN*, a method for learning behaviours directly in the parameterised action space. This leverages the distinct nature of the action space and is the current state-of-the-art algorithm on 2D robot soccer and King of Glory, a multiplayer online battle arena game. However, the formulation of the approach is flawed due to the dependence of the discrete action values on all action-parameters, not only those associated with each action. In this paper, we show how the above issue leads to suboptimal decision-making. We then introduce a novel multi-pass method to separate action-parameters, and demonstrate that the resulting algorithm---\MPDQN*---outperforms existing methods on the Platform, Robot Soccer Goal, and Half Field Offense domains.

\section{Background}

Parameterised action spaces \citep{masson2016} consist of a set of discrete actions, $\mathcal{A}_d= [K]=\{k_1,k_2,... ,k_K\}$, where each $k$ has a corresponding continuous action-parameter \mbox{$x_k \in \mathcal{X}_k \subseteq \Real^{m_k}$} with dimensionality $m_k$. This can be written as
\begin{equation}\label{eq:parameterised_action_spaces}
\mathcal{A} = \bigcup_{k \in [K]} \{a_k = (k, x_k) | x_k \in \mathcal{X}_k\}.
\end{equation}
We consider environments modelled as a Parameterised Action Markov Decision Process (PAMDP) \cite{masson2016}. For a PAMDP \mbox{$M = (\mathcal{S},\mathcal{A},P,R,\gamma)$}: $\mathcal{S}$ is the set of all states, $\mathcal{A}$ is the parameterised action space, $P(s'|s,k,x_k)$ is the Markov state transition probability function, $R(s,k,x_k,s')$ is the reward function, and $\gamma \in [0,1)$ is the future reward discount factor. An action policy $\pi: \mathcal{S} \rightarrow \mathcal{A}$ maps states to actions, typically with the aim of maximising Q-values $Q(s,a)$, which give the expected discounted return of executing action $a$  in state $s$ and following the current policy thereafter. 

The \QPAMDP* algorithm \citep{masson2016} alternates between learning a discrete action policy with fixed action-parameters using Sarsa($\lambda$) \citep{sutton1998} with the Fourier basis \citep{konidaris2011} and optimising the continuous action-parameters using episodic Natural Actor Critic (eNAC) \citep{peters2005} while the discrete action policy is kept fixed.
\citet{hausknecht2016} apply artificial neural networks and the Deep Deterministic Policy Gradients (DDPG) algorithm \citep{lillicrap2015} to parameterised action spaces by treating both the discrete actions and their action-parameters as a joint continuous action vector. This can be seen as relaxing the parameterised action space (Equation~\ref{eq:parameterised_action_spaces}) 
into a continuous one:
\begin{equation}
\mathcal{A} = \{(f_{1:K}, x_{1:K}) | f_k \in \Real, x_k \in \mathcal{X}_k \forall k \in [K]	\},
\end{equation}
where $f_1, f_2, \dots, f_K$ are continuous values in $[-1,1]$. An $\epsilon$-greedy or softmax policy is then used to select discrete actions. However, not only does this fail to exploit the disjoint nature of different parameterised actions, but optimising over the joint action and action-parameter space can result in premature convergence to suboptimal policies, as occurred in experiments by \citet{masson2016}. We henceforth refer to the algorithm used by \citeauthor{hausknecht2016} as \PADDPG*. 

\begin{figure}
	\centering
    \includegraphics[height=6cm]{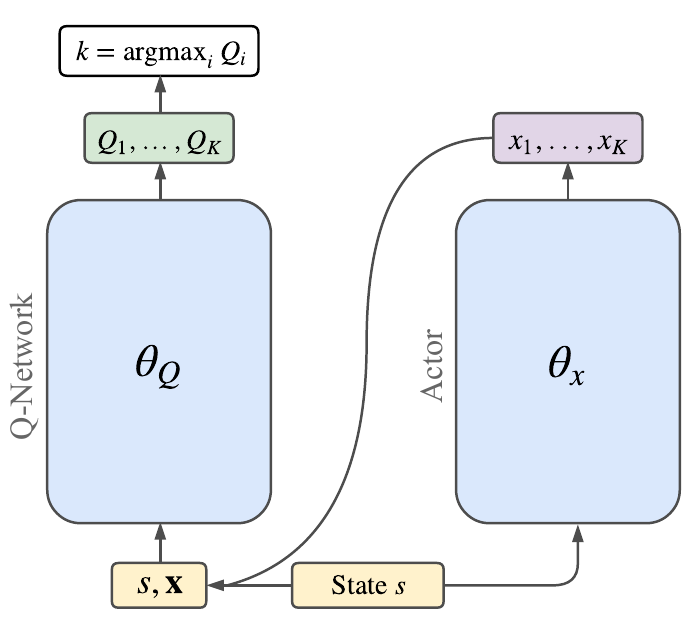}
	\caption[The \PDQN* network architecture]{The \PDQN* network architecture~\protect\citep{xiong2018}. Note that the joint action-parameter vector $\mathbf{x}$ is fed into the Q-network.}
	\label{fig:pdqn}
\end{figure}

\subsection{Parameterised Deep Q-Networks}
Unlike previous approaches, \citet{xiong2018} introduce a method that operates in the parameterised action space directly by combining DQN and DDPG. Their \PDQN* algorithm achieves state-of-the-art performance using a Q-network to approximate Q-values used for discrete action selection, in addition to providing critic gradients for an actor network that determines the continuous action-parameter values for all actions.
By framing the problem as a PAMDP directly, rather than alternating between discrete and continuous action MDPs as with \QPAMDP*, or using a joint continuous action MDP as with \PADDPG*, \PDQN* necessitates a change to the Bellman equation to incorporate continuous action-parameters:
\begin{equation}
Q(s, k, x_k) = \mathop{\Expec}_{r,s'}\Big[r + \gamma \max_{k'}\sup_{x_{k'}\in\mathcal{X}_{k'}}Q(s',k',x_{k'})\Big|s,k,x_k\Big].%
\end{equation}
To avoid the computationally intractable calculation of the supremum over $\mathcal{X}_k$, \citet{xiong2018} state that when the $Q$ function is fixed, one can view $\text{argsup}_{x_k\in\mathcal{X}_k}Q(s,k,x_k)$ as a function $x_k^Q : S \to \mathcal{X}_k$ for any state $s\in S$ and $k \in [K]$. This allows the Bellman equation to be rewritten as:
\begin{equation}\label{eq:parameterised_bellman}
Q(s, k, x_k) = \mathop{\Expec}_{r,s'}\left[r + \gamma \max_{k'}Q(s',k',x_{k'}^Q(s'))\Big|s,k,x_k\right].
\end{equation}
\PDQN* uses a deep neural network with parameters $\theta_Q$ to represent $Q(s,k,x_k;\theta_Q)$, and a second deterministic actor network with parameters $\theta_x$ to represent the action-parameter policy $x_k(s;\theta_x) : S \to \mathcal{X}_k$, an approximation of $x_k^Q(s)$. With this formulation it is easy to apply the standard DQN approach of minimising the mean-squared Bellman error to update the Q-network using minibatches sampled from replay memory $D$ \cite{mnih2015}, replacing $a$ with $(k, x_k)$:
\begin{equation}\label{eq:pdqn_qnetwork_loss}
L_Q(\theta_Q) = \mathop{\Expec}_{(s,k,x_k,r,s')\sim D}\Big[\frac{1}{2}\big(y - Q(s,k,x_k;\theta_Q)\big)^2\Big],
\end{equation}
where $y = r + \gamma\max_{k' \in [K]} Q(s',k', x_{k'}(s';\theta_x); \theta_Q)$ is the update target derived from Equation~\eqref{eq:parameterised_bellman}. Then, the loss for the actor network in \PDQN* is given by the negative sum of Q-values:
\begin{equation}\label{eq:action_parameter_loss}
L_x(\theta_x) = \mathop{\Expec}_{s \sim D}\Bigg[ - \sum_{k=1}^K Q\big(s,k,x_k(s;\theta_x);\theta_Q\big) \Bigg].
\end{equation}
Although this choice of loss function was not motivated by \citet{xiong2018}, it resembles the deterministic policy gradient loss used by \PADDPG* where a scalar critic value is used over all action-parameters \citep{hausknecht2016}. 
During updates, the estimated Q-values are backpropagated through the critic to the actor, producing gradients indicating how the action-parameters should be updated to increase the Q-values.

\section{Problems with Joint Action-Parameters}
\begin{figure*}[ht!]
	\begin{subfigure}[t]{0.32\textwidth}
		\centering
		\includegraphics[width=\textwidth]{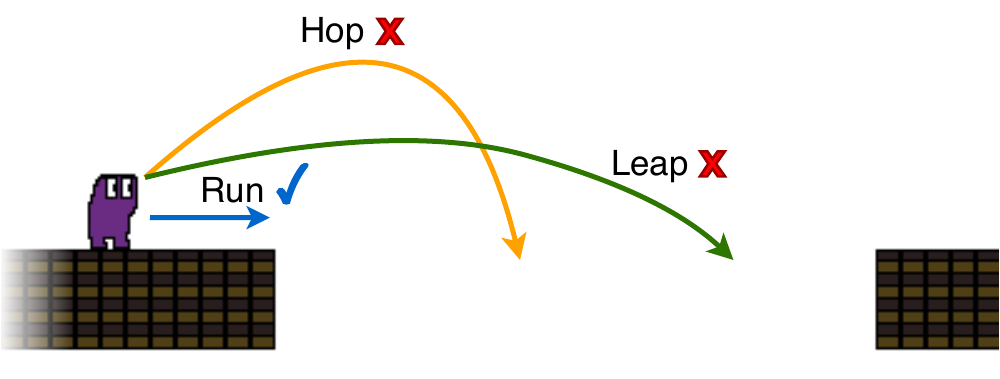}
		\caption{Agent in the Platform domain.}
		\label{fig:platform_state}
	\end{subfigure}%
	\hspace{0.1cm}
	\begin{subfigure}[t]{0.67\textwidth}
		\centering
		\includegraphics[width=\textwidth]{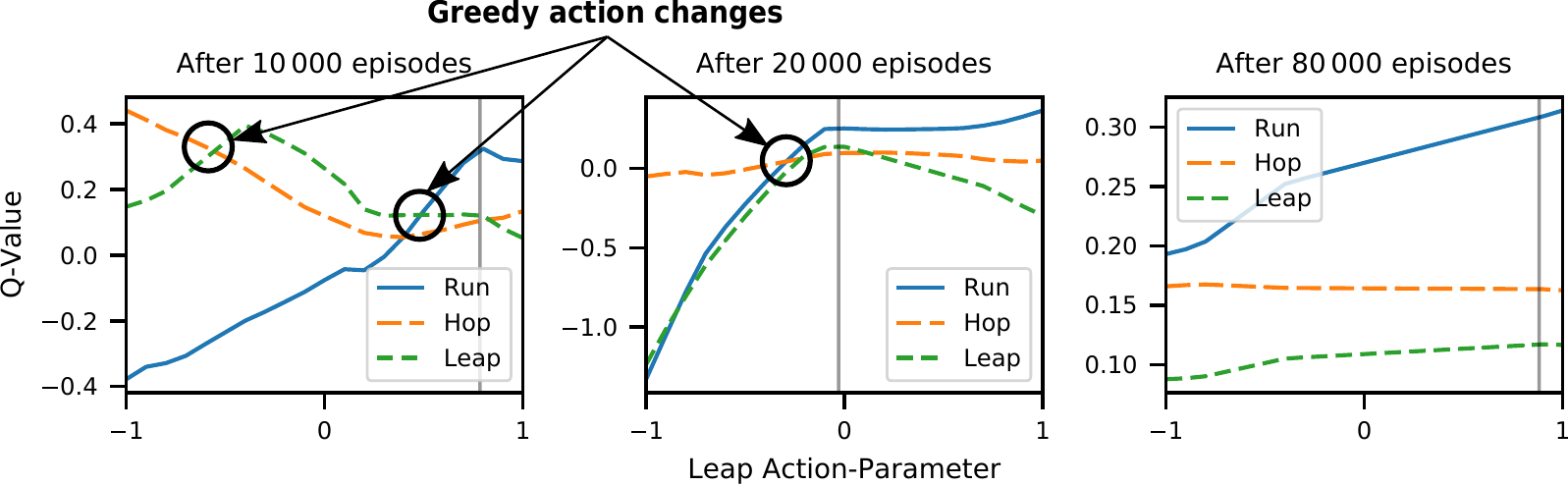}
		\caption{Predicted Q-values versus leap action-parameter. Vertical lines indicate the leap value actually chosen. Ideally, the run and hop Q-values should remain constant since their action-parameters are fixed.}
		\label{fig:platform_state_qvalues}
	\end{subfigure}
	\caption[Example of sensitivity to unrelated action-parameters affecting discrete action selection]{Example of dependence on unrelated action-parameters affecting discrete action selection on the Platform domain. Three parameterised actions are available: \emph{run}, \emph{hop}, and \emph{leap}. In a particular state (\subref{fig:platform_state}), the optimal action is to run forward to be able to traverse a gap, while choosing to leap would cause the agent to fall and die. The Q-value of the leap action should change with its action-parameter, but (\subref{fig:platform_state_qvalues}) shows that varying the leap action-parameter while the others are kept fixed changes the Q-values predicted by \PDQN* for \emph{all} actions. Near the start of training, this can alter the discrete policy such that a suboptimal action is chosen. After $80\,000$ episodes, \PDQN* correctly learns to choose the optimal action regardless of the unrelated leap action-parameter, although the other Q-values still vary.}
	\label{fig:example_sensitivity}
\end{figure*}
The \PDQN* architecture inputs the joint action-parameter vector over all actions to the Q-network, as illustrated in Figure~\ref{fig:pdqn}. This was pointed out by \citet{xiong2018} but they did not discuss it further. While this may seem like an inconsequential implementation detail, 
it 
changes the formulation of the Bellman equation used for parameterised actions (Equation~\ref{eq:parameterised_bellman}) since each Q-value is a function of the joint action-parameter vector \mbox{$\mathbf{x} = \left( x_{1}, \ldots, x_K\right)$}, rather than only the action-parameter $x_k$ corresponding to the associated action:
\begin{equation}\label{eq:parameterised_bellman_joint}
Q(s, k, \mathbf{x}) = \mathop{\Expec}_{r,s'}\Big[r + \gamma \max_{k'}Q(s',k',\mathbf{x}^Q(s'))\Big|s,k,\mathbf{x}\Big].
\end{equation}
This in turn affects both the updates to the Q-values and the action-parameters. Firstly, we consider the effect on the action-parameter loss, specifically that each Q-value produces gradients for all action-parameters. Consider for demonstration purposes the action-parameter loss (Equation~\ref{eq:action_parameter_loss}) over a single sample with state $s$:
\begin{equation}
L_x(\theta_x) = - \sum_{k=1}^K Q\big(s,k,\mathbf{x}(s;\theta_x);\theta_Q\big).
\end{equation}
The policy gradient is then given by:
\begin{equation}
\nabla_{\theta_x}\mathbf{x}(s; \theta_x) = - \sum_{k=1}^K \nabla_\mathbf{x} Q\big(s,k,\mathbf{x}(s;\theta_x);\theta_Q\big) \nabla_{\theta_x}\mathbf{x}(s; \theta_x).
\end{equation}
Expanding the gradients with respect to the action-parameters gives
\begin{equation}
\resizebox{\hsize}{!}{%
$\nabla_\mathbf{x}Q = \begin{pmatrix}
\frac{\partial Q_1}{\partial x_1} + \frac{\partial Q_2}{\partial x_1} + \cdots + \frac{\partial Q_K}{\partial x_1},  & \cdots & ,\frac{\partial Q_1}{\partial x_K} + \cdots + \frac{\partial Q_K}{\partial x_K} 
\end{pmatrix},$%
}
\end{equation}
where $Q_k = Q(s,k,\mathbf{x}(s;\theta_x); \theta_Q)$. Theoretically, if each Q-value were a function of just $x_k$ as the \PDQN* formulation intended, then \mbox{$\sfrac{\partial Q_k}{\partial x_j} = 0\ \forall k,j \in [K], j \ne k$} and $\nabla_\mathbf{x}Q$ simplifies to:
\begin{equation}
\nabla_\mathbf{x}Q = \begin{pmatrix}
\frac{\partial Q_1}{\partial x_1}, & \frac{\partial Q_2}{\partial x_2},  & \cdots & ,\frac{\partial Q_K}{\partial x_K} 
\end{pmatrix}.
\end{equation}
However this is not the case in \PDQN*, so the gradients with respect to other action-parameters $\sfrac{\partial Q_k}{\partial x_j}$ are not zero in general. This is a problem because each Q-value is updated only when its corresponding action is sampled, as per Equation~\ref{eq:pdqn_qnetwork_loss}, and thus has no information on what effect other action-parameters $x_{j}, j \ne k$ have on transitions or how they should be updated to maximise the expected return. They therefore produce what we term \emph{false gradients}. This effect may be mitigated by the summation over all Q-values in the action-parameter loss, since the gradients from each Q-value are summed and averaged over a minibatch.

The dependence of Q-values on all action-parameters also negatively affects the discrete action policy. Specifically, updating the continuous action-parameter policy of any action perturbs the Q-values of \emph{all} actions,  not just the one associated with that action-parameter. This can lead to the relative ordering of Q-values changing, which in turn can result in suboptimal greedy action selection. We demonstrate a situation where this occurs on the Platform domain in Figure~\ref{fig:example_sensitivity}.

\section{Multi-Pass Q-Networks}

The na\"ive solution to the problem of joint action-parameter inputs in \PDQN* would be to split the Q-network 
into separate networks for each discrete action. Then, one can input only the state and relevant action-parameter $x_k$ to the network corresponding to $Q_k$. However, this drastically increases the computational and space complexity of the algorithm due to the duplication of network parameters for each action. Furthermore, the loss of the shared feature representation between Q-values may be detrimental.

We therefore consider an alternative approach that does not involve architectural changes to the network structure of \PDQN*. While separating the action-parameters in a single forward pass of a single Q-network with fully connected layers is impossible, we can do so with multiple passes. We perform a forward pass once per action $k$ with the state $s$ and action-parameter vector $\mathbf{xe}_k$ as input, where $\mathbf{e}_k$ is the standard basis vector for dimension $k$. Thus $\mathbf{xe}_k = \left(0,\ldots,0,x_k,0,\ldots,0\right)$ is the joint action-parameter vector where each $x_j, j\ne k$ is set to zero. This causes all false gradients to be zero, $\sfrac{\partial Q_k}{\partial x_j} = 0$, and completely negates the impact of the network weights for unassociated action-parameters $x_j$ from the input layer, making $Q_k$ only depend on $x_k$. That is, 
\begin{equation}
Q\left(s,k,\mathbf{xe}_k\right) \approxeq Q\left(s,k,x_k\right).
\end{equation} 
Both  problems are therefore addressed without introducing any additional neural network parameters. We refer to this as the \emph{multi-pass Q-network} method, or \MPDQN*.

\begin{figure}[ht]
	\centering
	\includegraphics[height=8.2cm]{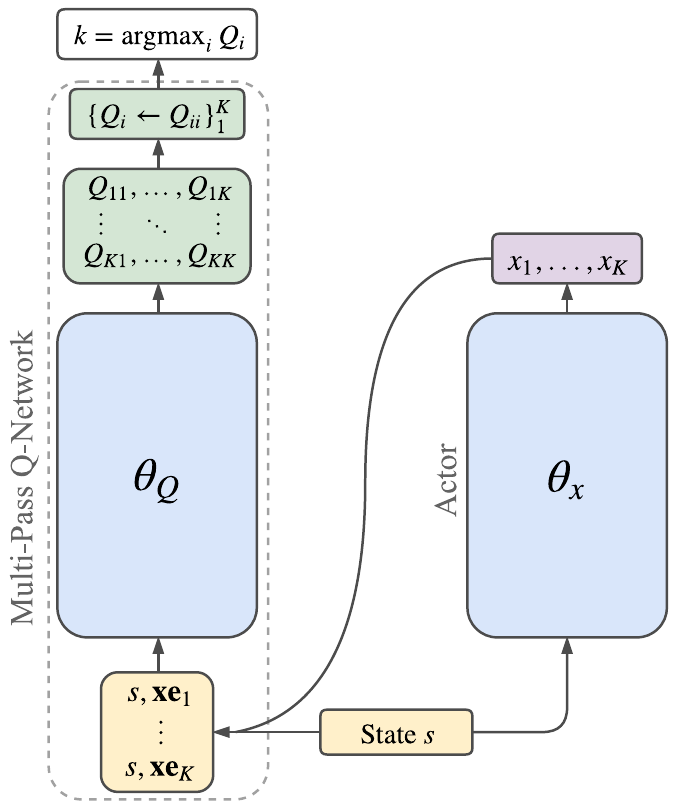}
	\caption{Illustration of the multi-pass Q-network architecture.}
	\label{fig:mpdqn}
\end{figure}

A total of $K$ forward passes are required to predict all Q-values instead of one. However, we can make use of the parallel minibatch processing capabilities of artificial neural networks, provided by libraries such as PyTorch and Tensorflow, to perform this in a single parallel pass, or \emph{multi-pass}. A multi-pass with $K$ actions is processed in the same manner as a minibatch of size $K$:
\begin{equation}
\begin{pmatrix}
Q\left( s,\cdotp,\mathbf{xe}_1; \theta_Q \right)\\
\vdots \\
Q\left( s,\cdotp,\mathbf{xe}_K; \theta_Q \right)
\end{pmatrix}=
\begin{pmatrix}
Q_{11} & Q_{12}  & \cdots & Q_{1K}\\
\vdots & \vdots  &\ddots  & \vdots \\
Q_{K1} & Q_{K2}  & \cdots & Q_{KK}
\end{pmatrix},
\end{equation}
where $Q_{ij}$ is the Q-value for action $j$ generated on the $i$\textsuperscript{th} pass where $x_i$ is non-zero. Only the diagonal elements $Q_{ii}$ are valid and used in the final output $Q_i \gets Q_{ii}$. This process is illustrated in Figure~\ref{fig:mpdqn}.

Compared to separate Q-networks, our multi-pass technique introduces a relatively minor amount of overhead during forward passes. Although minibatches for updates are similarly duplicated $K$ times, backward passes to accumulate gradients are not duplicated since only the diagonal elements $Q_{ii}$ are used in the loss function. The computational complexity of this overhead scales linearly with the number of actions and minibatch size during updates. Unlike separate Q-networks (and even when a larger Q-network with more hidden layers and neurons is used) if the number of actions does not change, then the overhead of multi-passes would be the same as with a smaller Q-network, provided the minibatch is of a reasonable size and can be processed in parallel.

\section{Experiments}

We compare the original \PDQN* algorithm with a single Q-network against our proposed multi-pass Q-network (\MPDQN*), as well as against separate Q-networks (\SPDQN*). We also compare against \QPAMDP* and \PADDPG*, the former state-of-the-art approaches on their respective domains. We are unable to use King of Glory as a benchmark domain as it is closed-source and proprietary.

Similar to \citet{mnih2015} and \citet{hausknecht2016}, we add target networks to \PDQN* to compute the update targets $y$ for stability. Soft updates (Polyak averaging) are used for the target networks. Adam \citep{kingma2014} with $\beta_1 = 0.9, \beta_2 =0.999$ is used to optimise the neural network parameters for \PDQN* and \PADDPG*. Layer weights are initialised following the strategy of \citet{he2015} with rectified linear unit (ReLU) activation functions. We employ the inverting gradients approach to bound action-parameters for both algorithms, as \citet{hausknecht2016} claim \PADDPG* is unable to learn without it on Half Field Offense. Action-parameters are scaled to $[-1,1]$, as we found this increased performance for all algorithms. 

We perform a hyperparameter grid search for Platform and Robot Soccer Goal over: the network learning rates $\alpha_Q, \alpha_x \in \{10^{-1}, 10^{-2}, 10^{-3}, 10^{-4}, 10^{-5}\}~\text{s.t.}~\alpha_x \le \alpha_Q$; Polyak averaging factors $\tau_Q, \tau_x \in \{0.1, 0.01, 0.001\}~\text{s.t.}~\tau_x \le \tau_Q$; minibatch size $B \in \{32, 64, 128\}$; and number of hidden layers and neurons in $\{(256, 128), (128, 64), (256), (128)\}$. The hidden layers are kept symmetric between the actor and critic networks as in previous works. Each combination is tested over $5$ random runs for \PDQN* and \PADDPG* separately on each domain. The same hyperparameters are used for \PDQN*, \SPDQN* and \MPDQN*. 

To keep the comparison with \PADDPG* fair, we do not use dueling networks \citep{wang2016} nor asynchronous parallel workers as \citet{xiong2018} used for \PDQN*. For each algorithm and domain, we train $30$ agents with unique random seeds and evaluate them without exploration for another $1000$ episodes. Our experiments are implemented in Python using PyTorch \citep{paszke2017} and OpenAI Gym \citep{openaigym}, and run on the following hardware: Intel Core i7-7700, 16GB DRAM, NVidia GTX 1060 GPU. Complete source code is available online.\footnote{\url{https://github.com/cycraig/MP-DQN}}

\subsection{Platform}
The Platform domain \citep{masson2016} has three actions---run, hop, and leap---each with a continuous action-parameter to control horizontal displacement. The agent has to hop over enemies and leap across gaps between platforms to reach the goal state. The agent dies if it touches an enemy or falls into a gap. A $9$-dimensional state space gives the position and velocity of the agent and local enemy along with features of the current platform such as length. 

We train agents on this domain for $80\,000$ episodes, using the same hyperparameters for \QPAMDP* as \citet{masson2016}, except we reduce the learning rate for eNAC ($\alpha_{\text{eNAC}}$) to $0.1$, and exploration noise variance ($\sigma$) to $0.0001$, to account for the scaled action-parameters. For \PDQN*, shallow networks with one hidden layer $(128)$ were found to perform best with $\alpha_Q = 10^{-3}$, $\alpha_x = 10^{-4}$, $\tau_Q = 0.1$, $\tau_x = 0.001$, and $B = 128$. \PADDPG* uses two hidden layers $(256,128)$ with $\alpha_Q = 10^{-3}$, $\alpha_\mu = 10^{-4}$, $\tau_Q = 0.01$, $\tau_\mu = 0.01$, and $B = 32$. A replay memory size of $10\,000$ samples is used for both algorithms, update gradients are clipped at $10$, and $\gamma = 0.9$.

We introduce a \emph{passthrough layer} to the actor networks of \PDQN* and \PADDPG* to initialise their action-parameter policies to the same linear combination of state variables that \citet{masson2016} use to initialise the \QPAMDP* policy. The weights of the passthrough layer are kept fixed to avoid instability; this does not reduce the range of action-parameters available as the output of the actor network compensates before inverting gradients are applied. We use an $\epsilon$-greedy discrete action policy with additive Ornstein-Uhlenbeck noise for action-parameter exploration, similar to \citet{lillicrap2015}, which we found gives slightly better performance than Gaussian noise.

\subsection{Robot Soccer Goal}
The Robot Soccer Goal domain \citep{masson2016} is a simplification of RoboCup 2D \citep{kitano1997} in which an agent has to score a goal past a keeper that tries to intercept the ball. The three parameterised actions---kick-to, shoot-goal-left, and shoot-goal-right---are all related to kicking the ball, which the agent automatically approaches between actions until close enough to kick again. The state space consists of $14$ continuous features describing the position, velocity, and orientation of the agent and keeper, and the ball's position and distance to the keeper and goal.

\begin{figure}[ht!]
	\begin{subfigure}[t]{\linewidth}
		\centering
		\includegraphics[width=0.95\textwidth]{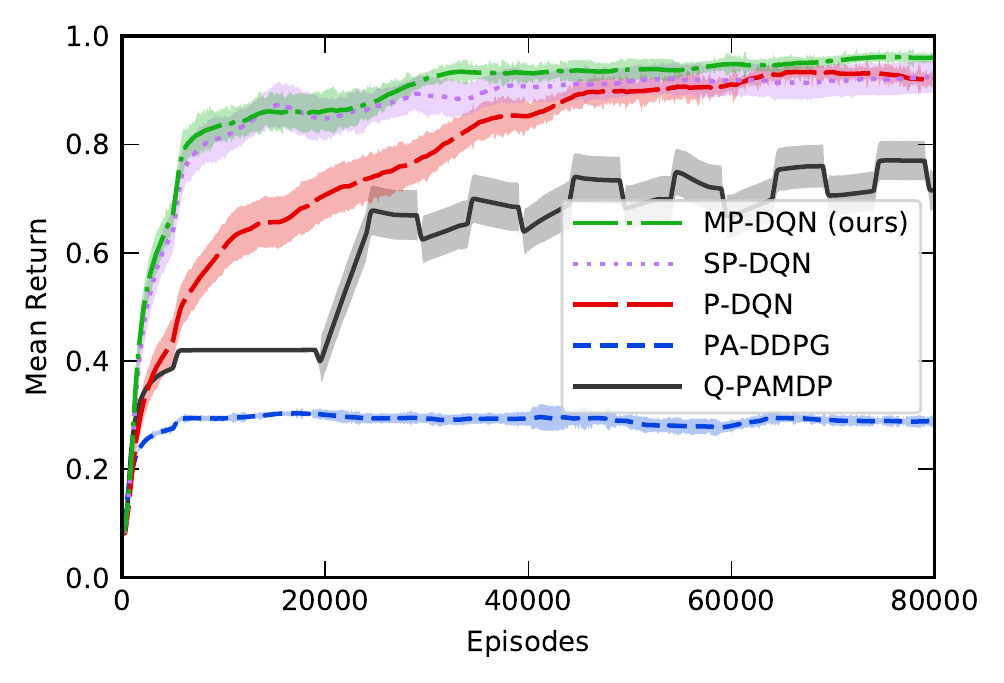}
		\caption{Platform}
		\label{fig:learning_curves_platform}
	\end{subfigure}
	\vspace*{0.15cm}
	\begin{subfigure}[t]{\linewidth}
		\centering
		\includegraphics[width=0.95\textwidth]{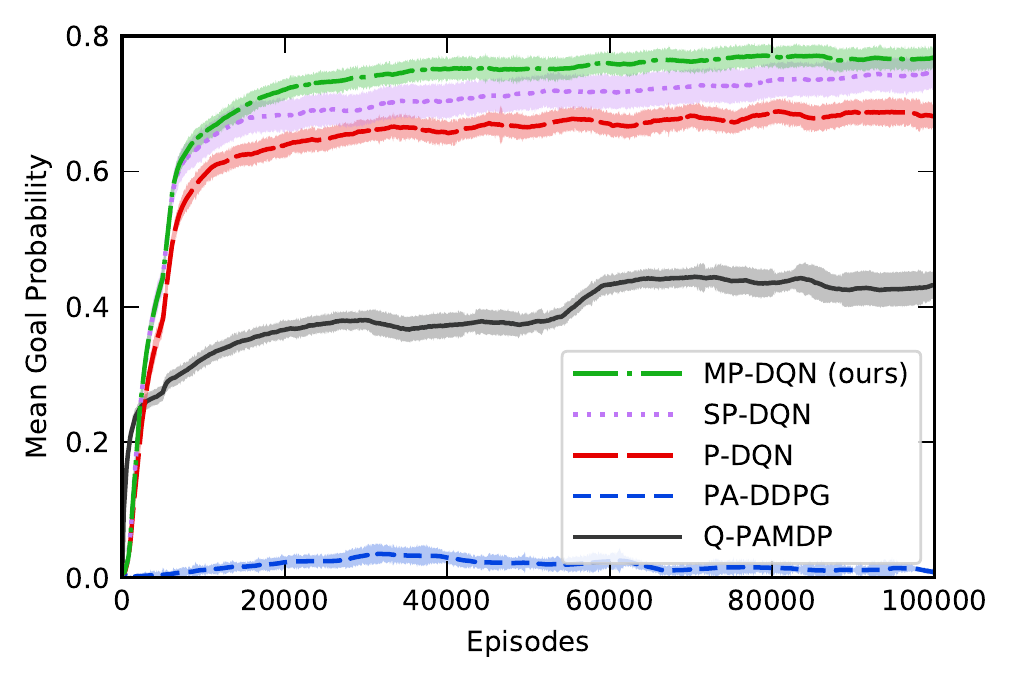}
		\caption{Robot Soccer Goal}
		\label{fig:learning_curves_robot_soccer_goal}
	\end{subfigure}
	\vspace*{0.15cm}
	\begin{subfigure}[t]{\linewidth}
		\centering
		\includegraphics[width=0.95\textwidth]{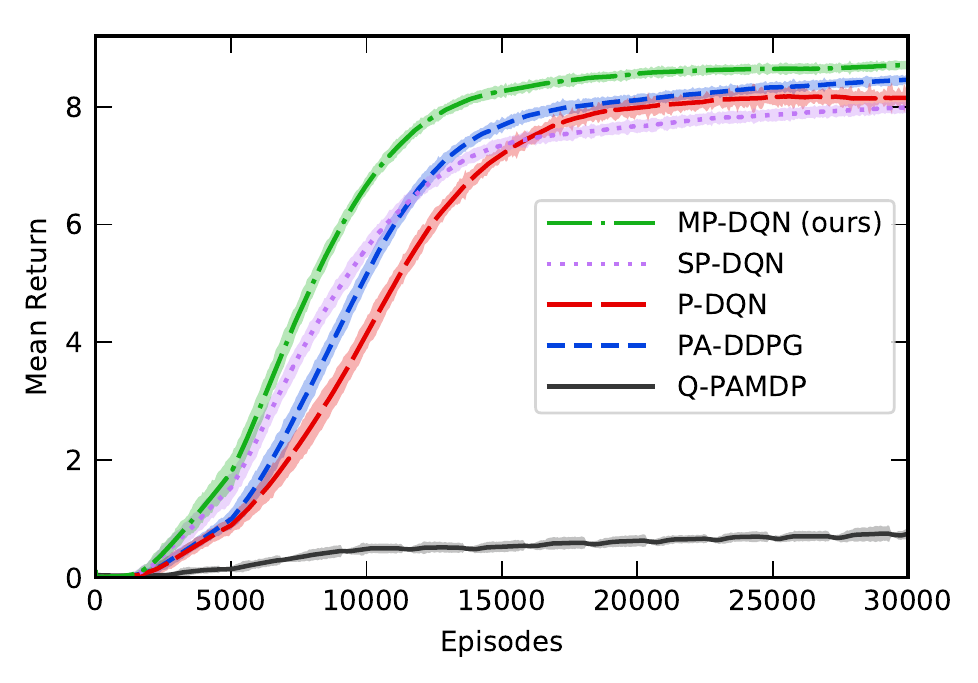}
		\caption{Half Field Offense}
		\label{fig:learning_curves_hfo}
	\end{subfigure}
	\caption{Learning curves on Platform (\subref{fig:learning_curves_platform}), Robot Soccer Goal (\subref{fig:learning_curves_robot_soccer_goal}), and Half Field Offense (\subref{fig:learning_curves_hfo}). The running average scores---episodic return for Platform and HFO, and goal scoring probability for Robot Soccer Goal---are smoothed over $5000$ episodes and include random exploration, higher is better. Shaded areas represent standard error of the running averages over the different agents. The oscillating behaviour of \QPAMDP* on Platform is a result of re-exploration between the alternating optimisation steps. \MPDQN* clearly performs best overall. }
	\label{fig:learning_curves}
\end{figure}

Training consisted of $100\,000$ episodes, using the same hyperparameters for \QPAMDP* as \citet{masson2016} except we set $\alpha_{\text{eNAC}} = 0.06$ and $\sigma = 0.0001$. \PDQN* uses a single hidden layer $(256)$, with $\alpha_Q = 10^{-3}$, $\alpha_x = 10^{-5}$, $\tau_Q = 0.1$, $\tau_x = 0.001$, and $B = 128$. Two hidden layers $(128,64)$ are used for \PADDPG*, with $\alpha_Q = 10^{-4}$, $\alpha_\mu = 10^{-5}$, $\tau_Q = 0.01$, $\tau_\mu = 0.01$, and $B = 64$. Both algorithms use a replay memory size of $20\,000$, $\gamma = 0.95$, gradients clipping at $1$, and the same action-parameter policy initialisation as \QPAMDP* with additive Ornstein-Uhlenbeck noise.

\begin{table*}[ht!]
	\centering
	\begin{tabular}{@{}
			l
			S[table-format=1.3(4),detect-weight,mode=text]
			S[table-format=1.3(4),detect-weight,mode=text]
			S[table-format=1.3(4),detect-weight,mode=text]
			S[table-format=3(2),detect-weight,mode=text]
			@{}}
		\toprule
		& \multicolumn{1}{c}{{\textbf{Platform}}} & \multicolumn{1}{c}{\textbf{Robot Soccer Goal}} & \multicolumn{2}{c}{\textbf{Half Field Offense}} \\
		& {\textbf{Return}} & {\textbf{P(Goal)}} & {\textbf{P(Goal)}} & {\textbf{Avg. Steps to Goal}} \\ \midrule
		\QPAMDP* & 0.789 \pm 0.188 & 0.452 \pm 0.093 &{$0 \pm 0$} & {n/a}  \\
		\PADDPG* & 0.284 \pm 0.061 & 0.006 \pm 0.020 & 0.875\pm 0.182 & \B 95 \pm 7 \\
		\PDQN*   & 0.964 \pm 0.068 & 0.701 \pm 0.078 & 0.883\pm 0.085 & 111 \pm 11 \\
		\SPDQN* & 0.941 \pm 0.164 & 0.752 \pm 0.131 & 0.718 \pm 0.131 & 99 \pm 7 \\
		\MPDQN*  & \B 0.987 \pm 0.039 & \B 0.789 \pm 0.070 & \B 0.913\pm 0.070 & 99 \pm 12 \\ 
		\cdashlinelr{1-5}
		\PADDPG*\footnotemark[2] &{-}&{-}& 0.923 \pm 0.073 & 112 \pm 5 \\
		Async. \PDQN*\footnotemark[3] &{-}&{-}& 0.989 \pm 0.006 & 81 \pm 3 \\ \bottomrule 
	\end{tabular}
	\caption[Table]{Mean evaluation scores over $30$ random runs for each algorithm, averaged over $1000$ episodes after training with no random exploration. We include previously published results from \citet{hausknecht2016} and \citet{xiong2018} on HFO, although they are not directly comparable with ours as we use a longer training period and have a much larger sample size of agents---$30$ versus $7$ and $9$ respectively---and asynchronous \PDQN* uses $24$ parallel workers to implement $n$-step returns rather than the mixing strategy we use.}
	\label{tab:results_comparison}
\end{table*}

\subsection{Half Field Offense}
The third and final domain, Half Field Offense (HFO) \citep{hausknecht2016}, is also the most complex. It has $58$ state features and three parameterised actions available: dash, turn, and kick. Unlike Robot Soccer Goal, the agent must first learn to approach the ball and then kick it into the goals, although there is no keeper in this task. 

We use $30\,000$ episodes for training on HFO. This is more than the $20\,000$ episodes (or roughly $3$ million transitions) used by \citet{hausknecht2016} and \citet{xiong2018} so that ample opportunity is given for the algorithms to converge in order to fairly evaluate the final policy performance. We use the same network structure as previous works with hidden layers of $(256,128,64)$ neurons for \PDQN* and $(1024,512,256,128)$ neurons for \PADDPG*. The leaky ReLU activation function with negative slope $0.01$ is used on HFO because of these deeper networks. \citet{xiong2018} use $24$ asynchronous parallel workers for $n$-step returns on HFO. For fair comparison and due to the lack of sufficient hardware, we instead use mixed $n$-step return targets \cite{hausknecht2016betamixing} with a mixing ratio of $\beta = 0.25$ for both \PDQN* and \PADDPG*, as this technique does not require multiple workers. The $\beta$ value was selected after a search over $\beta \in \{0, 0.25, 0.5, 0.75, 1\}$. We otherwise use the same hyperparameters as \citet{hausknecht2016betamixing} apart from the network learning rates: $\alpha_Q = 10^{-3}$, $\alpha_x = 10^{-5}$ for \PDQN* and $\alpha_Q = 10^{-3}$, $\alpha_\mu = 10^{-3}$ for \PADDPG*. In the absence of an initial action-parameter policy, we use the same $\epsilon$-greedy with uniform random action-parameter exploration strategy as the original authors. In general we kept as many factors consistent between the two algorithms as possible for a fair comparison.

We select $10$ of the most relevant state features for \QPAMDP* to avoid intractable Fourier basis calculations. These features include: player orientation, stamina, proximity to ball, ball angle, ball-kickable, goal centre position, and goal centre proximity. Even with this reduced selection, we found at most a Fourier basis of order $2$ could be used. We use an adaptive step-size \citep{dabney2012} for Sarsa($\lambda$) with an eNAC learning rate of $0.2$. The Q-PAMDP agent initially learns with Sarsa($\lambda$) for a period of $1000$ episodes before alternating between $\kappa = 50$ eNAC updates of $25$ rollouts each, and $1000$ episodes of discrete action re-exploration.
\footnotetext[2]{Average over $7$ runs \citep{hausknecht2016}.}
\footnotetext[3]{Average over $9$ runs with $24$ workers \citep{xiong2018}.}

\section{Results}

The resulting learning curves of \MPDQN*, \SPDQN*, \PDQN*, \PADDPG*, and \QPAMDP* on the three parameterised action benchmark domains are shown in Figure~\ref{fig:learning_curves}, with mean evaluation scores detailed in Table~\ref{tab:results_comparison}.  

Our results show that \MPDQN* learns significantly faster than baseline \PDQN* with joint action-parameter inputs and achieves the highest mean evaluation scores across all three domains. \SPDQN* similarly shows better performance than \PDQN* on Platform and Robot Soccer Goal but to a slightly lesser extent than \MPDQN*. Notably, \SPDQN* exhibits fast initial learning on HFO but plateaus at a lower performance level than \PDQN*. This is likely due to the aforementioned lack of a shared feature representation between the separate Q-networks and the duplicate network parameters which require more updates to optimise.

In general, we observe that \PDQN* and its variants outperform \QPAMDP* on Platform and Robot Soccer Goal, while \PADDPG* consistently converges prematurely to suboptimal policies. \citet{wei2018} observe similar behaviour for \PADDPG* on Platform. This highlights the problem with updating the action and action-parameter policies simultaneously and was also observed when using eNAC for direct policy search on Platform \citep{masson2016}. On HFO, \QPAMDP* fails to learn to score any goals---likely due to its reduced feature space and use of linear function approximation rather than neural networks. Unexpectedly, baseline \PDQN* appears to learn slower than \PADDPG* on HFO. This suggests that the dueling networks and asynchronous parallel workers used by \citet{xiong2018} were major factors improving \PDQN* in their comparisons.



\section{Related Work}

Many recent deep RL approaches follow the strategy of collapsing the parameterised action space into a continuous one. \citet{hussein2018} present a deep imitation learning approach for scoring goals on HFO using long-short-term-memory networks with a joint action and action-parameter policy. \citet{agarwa2018} introduces skills for multi-goal parameterised action space environments to achieve multiple related goals; they demonstrate success on robotic manipulation tasks by combining \PADDPG* with hindsight experience replay and their skill library.

One can alternatively view parameterised actions as a 2-level hierarchy: \citet{klimek2017} use this approach to learn a reach-and-grip task using a single network to represent a distribution over macro (discrete) actions and their lower-level action-parameters. The work most relevant to this paper is by \citet{wei2018}, who introduce a parameterised action version of TRPO (PATRPO). They also take a hierarchical approach but instead condition the action-parameter policy on the discrete action chosen to avoid predicting all action-parameters at once. While their preliminary results show the method achieves good performance on Platform, we omit comparison with PATRPO as it fails to learn to
score goals on HFO.

\section{Conclusion}
We identified a significant problem with the \PDQN* algorithm for parametrised action spaces: the dependence of its Q-values on all action-parameters causes false gradients and can lead to suboptimal action selection. We introduced a new algorithm, \MPDQN*, with separate action-parameter inputs which demonstrated superior performance over \PDQN* and former state-of-the-art techniques \QPAMDP* and \PADDPG*. We also found that \PADDPG* was unstable and converged to suboptimal policies on some domains. Our results suggest that future approaches should leverage the disjoint nature of parameterised action spaces and avoid simultaneous optimisation of the policies for discrete actions and continuous action-parameters.

\section*{Acknowledgments}
This work is based on the research supported in part by the National Research Foundation of South Africa (Grant Number: 113737).

\bibliographystyle{named}
\bibliography{references}

\begin{thebibliography}{}

\bibitem[\protect\citeauthoryear{Agarwal}{2018}]{agarwa2018}
Arpit Agarwal.
\newblock Deep reinforcement learning with skill library: Exploring with
  temporal abstractions and coarse approximate dynamics models.
\newblock Master's thesis, Carnegie Mellon University, Pittsburgh, PA, July
  2018.

\bibitem[\protect\citeauthoryear{Brockman \bgroup \em et al.\egroup
  }{2016}]{openaigym}
Greg Brockman, Vicki Cheung, Ludwig Pettersson, Jonas Schneider, John Schulman,
  Jie Tang, and Wojciech Zaremba.
\newblock Open{AI} gym.
\newblock {\em arXiv preprint arXiv:1606.01540}, 2016.

\bibitem[\protect\citeauthoryear{Dabney and Barto}{2012}]{dabney2012}
William Dabney and Andrew~G Barto.
\newblock Adaptive step-size for online temporal difference learning.
\newblock In {\em Proceedings of the Twenty-Sixth AAAI Conference on Artificial
  Intelligence}, 2012.

\bibitem[\protect\citeauthoryear{Hausknecht and Stone}{2016a}]{hausknecht2016}
Matthew Hausknecht and Peter Stone.
\newblock Deep reinforcement learning in parameterized action space.
\newblock In {\em Proceedings of the International Conference on Learning
  Representations}, 2016.

\bibitem[\protect\citeauthoryear{Hausknecht and
  Stone}{2016b}]{hausknecht2016betamixing}
Matthew Hausknecht and Peter Stone.
\newblock On-policy vs. off-policy updates for deep reinforcement learning.
\newblock In {\em Deep Reinforcement Learning: Frontiers and Challenges, IJCAI
  Workshop}, July 2016.

\bibitem[\protect\citeauthoryear{He \bgroup \em et al.\egroup }{2015}]{he2015}
Kaiming He, Xiangyu Zhang, Shaoqing Ren, and Jian Sun.
\newblock Delving deep into rectifiers: surpassing human-level performance on
  {ImageNet} classification.
\newblock In {\em Proceedings of the IEEE international conference on computer
  vision}, pages 1026--1034, 2015.

\bibitem[\protect\citeauthoryear{Hussein \bgroup \em et al.\egroup
  }{2018}]{hussein2018}
Ahmed Hussein, Eyad Elyan, and Chrisina Jayne.
\newblock Deep imitation learning with memory for {Robocup} soccer simulation.
\newblock In {\em Proceedings of the International Conference on Engineering
  Applications of Neural Networks}, pages 31--43. Springer, 2018.

\bibitem[\protect\citeauthoryear{Khamassi \bgroup \em et al.\egroup
  }{2017}]{khamassi2017}
Mehdi Khamassi, George Velentzas, Theodore Tsitsimis, and Costas Tzafestas.
\newblock Active exploration and parameterized reinforcement learning applied
  to a simulated human-robot interaction task.
\newblock In {\em Proceedings of the First IEEE International Conference on
  Robotic Computing}, pages 28--35. IEEE, 2017.

\bibitem[\protect\citeauthoryear{Kingma and Ba}{2014}]{kingma2014}
Diederik~P. Kingma and Jimmy~Lei Ba.
\newblock Adam: {A} method for stochastic optimization.
\newblock {\em arXiv preprint arXiv:1412.6980}, 2014.

\bibitem[\protect\citeauthoryear{Kitano \bgroup \em et al.\egroup
  }{1997}]{kitano1997}
Hiroaki Kitano, Minoru Asada, Yasuo Kuniyoshi, Itsuki Noda, Eiichi Osawa, and
  Hitoshi Matsubara.
\newblock Robocup: A challenge problem for {AI}.
\newblock {\em AI Magazine}, 18:73--85, 1997.

\bibitem[\protect\citeauthoryear{Klimek \bgroup \em et al.\egroup
  }{2017}]{klimek2017}
Maciej Klimek, Henryk Michalewski, and Piotr Miłoś.
\newblock Hierarchical reinforcement learning with parameters.
\newblock In {\em Conference on Robot Learning}, pages 301--313, 2017.

\bibitem[\protect\citeauthoryear{Konidaris \bgroup \em et al.\egroup
  }{2011}]{konidaris2011}
George~D. Konidaris, Sarah Osentoski, and Philip~S. Thomas.
\newblock Value function approximation in reinforcement learning using the
  {F}ourier basis.
\newblock In {\em Proceedings of the Twenty-Fifth Conference on Artificial
  Intelligence}, pages 380--385, August 2011.

\bibitem[\protect\citeauthoryear{Lillicrap \bgroup \em et al.\egroup
  }{2016}]{lillicrap2015}
Timothy~P Lillicrap, Jonathan~J Hunt, Alexander Pritzel, Nicolas Heess, Tom
  Erez, Yuval Tassa, David Silver, and Daan Wierstra.
\newblock Continuous control with deep reinforcement learning.
\newblock In {\em Proceedings of the International Conference on Learning
  Representations}, 2016.

\bibitem[\protect\citeauthoryear{Masson \bgroup \em et al.\egroup
  }{2016}]{masson2016}
Warwick Masson, Pravesh Ranchod, and George Konidaris.
\newblock Reinforcement learning with parameterized actions.
\newblock In {\em Proceedings of the Thirtieth AAAI Conference on Artificial
  Intelligence}, pages 1934--1940, 2016.

\bibitem[\protect\citeauthoryear{Mnih \bgroup \em et al.\egroup
  }{2015}]{mnih2015}
Volodymyr Mnih, Koray Kavukcuoglu, David Silver, Andrei~A Rusu, Joel Veness,
  Marc~G Bellemare, Alex Graves, Martin Riedmiller, Andreas~K Fidjeland, Georg
  Ostrovski, et~al.
\newblock Human-level control through deep reinforcement learning.
\newblock {\em Nature}, 518(7540):529, 2015.

\bibitem[\protect\citeauthoryear{Paszke \bgroup \em et al.\egroup
  }{2017}]{paszke2017}
Adam Paszke, Sam Gross, Soumith Chintala, Gregory Chanan, Edward Yang, Zachary
  DeVito, Zeming Lin, Alban Desmaison, Luca Antiga, and Adam Lerer.
\newblock Automatic differentiation in {PyTorch}.
\newblock In {\em NIPS 2017 Autodiff Workshop: The Future of Gradient-based
  Machine Learning Software and Techniques}, 2017.

\bibitem[\protect\citeauthoryear{Peng \bgroup \em et al.\egroup
  }{2016}]{peng2016}
Xue~Bin Peng, Glen Berseth, and Michiel Van~de Panne.
\newblock Terrain-adaptive locomotion skills using deep reinforcement learning.
\newblock {\em ACM Transactions on Graphics}, 35(4):81:1--81:12, 2016.

\bibitem[\protect\citeauthoryear{Peters and Schaal}{2008}]{peters2005}
Jan Peters and Stefan Schaal.
\newblock Natural actor-critic.
\newblock {\em Neurocomputing}, 71(7-9):1180--1190, 2008.

\bibitem[\protect\citeauthoryear{Schulman \bgroup \em et al.\egroup
  }{2015}]{schulman2015}
John Schulman, Sergey Levine, Pieter Abbeel, Michael~I Jordan, and Philipp
  Moritz.
\newblock Trust region policy optimization.
\newblock In {\em International Conference of Machine Learning}, volume~37,
  pages 1889--1897, 2015.

\bibitem[\protect\citeauthoryear{Sutton and Barto}{1998}]{sutton1998}
Richard~S. Sutton and Andrew~G. Barto.
\newblock {\em Reinforcement Learning: An Introduction}.
\newblock MIT Press, Cambridge, MA, USA, 1998.

\bibitem[\protect\citeauthoryear{Wang \bgroup \em et al.\egroup
  }{2016}]{wang2016}
Ziyu Wang, Tom Schaul, Matteo Hessel, Hado Van~Hasselt, Marc Lanctot, and Nando
  De~Freitas.
\newblock Dueling network architectures for deep reinforcement learning.
\newblock In {\em Proceedings of the 33rd International Conference on
  International Conference on Machine Learning - Volume 48}, pages 1995--2003,
  2016.

\bibitem[\protect\citeauthoryear{Wei \bgroup \em et al.\egroup
  }{2018}]{wei2018}
Ermo Wei, Drew Wicke, and Sean Luke.
\newblock Hierarchical approaches for reinforcement learning in parameterized
  action space.
\newblock In {\em 2018 AAAI Spring Symposium Series}, 2018.

\bibitem[\protect\citeauthoryear{Xiong \bgroup \em et al.\egroup
  }{2018}]{xiong2018}
Jiechao Xiong, Qing Wang, Zhuoran Yang, Peng Sun, Lei Han, Yang Zheng, Haobo
  Fu, Tong Zhang, Ji~Liu, and Han Liu.
\newblock Parametrized deep {Q}-networks learning: Reinforcement learning with
  discrete-continuous hybrid action space.
\newblock {\em arXiv preprint arXiv:1810.06394}, 2018.

\end{thebibliography}

\end{document}